\documentclass[a4paper]{article}
\usepackage{graphicx}
\usepackage{subfigure}
\usepackage{twocolceurws}

\title{Sparse Adversarial Attack to Object Detection}

\author{
Jiayu Bao\\ Department of Electronic Engineering Department, Tsinghua University\\
                Beijing, China 100084 \\ bjy19@mails.tsinghua.edu.cn
}

\institution{}

\begin{document}
\maketitle

\begin{abstract}
Adversarial examples have gained tons of attention in recent years. Many adversarial attacks have been proposed to attack image classifiers, but few work shift attention to object detectors. In this paper, we propose Sparse Adversarial Attack (SAA) which enables adversaries to perform effective evasion attack on detectors with bounded \emph{l$_{0}$} norm perturbation. We select the fragile position of the image and designed evasion loss function for the task. Experiment results on YOLOv4 and FasterRCNN reveal the effectiveness of our method. In addition, our SAA shows great transferability across different detectors in the black-box attack setting. Codes are available at \emph{https://github.com/THUrssq/Tianchi04}.
\end{abstract}

\section{Introduction}

Deep neural networks have achieved remarkable success in computer vision tasks like image classification, object detection and instance segmentation. Deep object detectors can be divided into one-stage detectors and two-stage detectors. One-stage detectors like YOLOv4 \cite{bochkovskiy2020yolov4} and SSD \cite{liu2016ssd} take classification and regression as a single step. Hence they are usually faster than two-stage detectors. While two-stage detectors with region proposal process often have better performance on accuracy than one-stage detectors. 

However, the finding that deep models are vulnerable to adversarial examples \cite{szegedy2013intriguing} poses great concerns for the security of deep models. In computer vision field, adversarial examples are perturbed images which are designed purposely to fool the deep neural models. Many adversarial attack methods have been proposed to study the robustness of deep models but most of them \cite{madry2017towards} \cite{carlini2017towards} \cite{brown2017adversarial} focus on image classifiers rather than the more widely used object detectors. We believe adversarial attacks play a significant role in improving the robustness of deep neural models. Existing adversarial attacks usually change the category classification of an image or an object \cite{xie2017adversarial} \cite{huang2020universal}, which is reasonable for classifiers but deficient for detectors.

\begin{figure}[htbp]
	\centering
	\vspace{0.1cm}
	\includegraphics[width=0.98\linewidth]{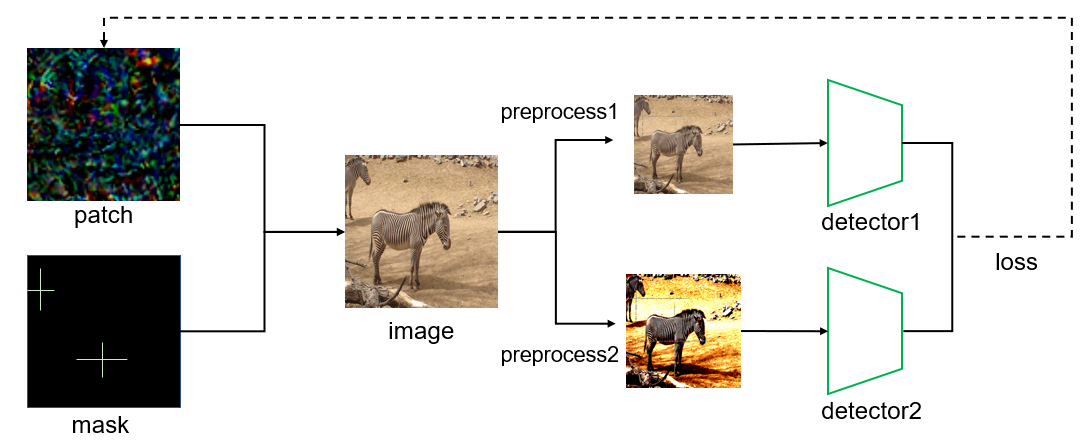}
	\caption{Framework of sparse adversarial attack.}
\end{figure}

Due to the complexity of object detection task, adversarial attacks on object detectors can be more diversified than that of classifiers. In this paper, we propose a pixel efficient evasion attack method SAA for object detectors. Concretely, the perturbation generated by our SAA can make all the objects in the image to evade detections of target detectors. We design different loss functions according to the model we attack and choose pivotal positions in the image to make our adversarial perturbation sparse yet powerful.

Experiment results on two state-of-the-art white-box detectors and two unknown black-box detectors show that our SAA is a pixel efficient adversarial attack with good transferability. Our method outperforms many other methods in the same task and get 4th in \emph{AIC Phase IVCIKM-2020: Adversarial Challenge on Object Detection} competition. The contributions of this work are as follows:
\begin{itemize}
	\item We propose sparse adversarial attack (SAA) method to perform evasion attack on object detectors.
	\item We design powerful evasion loss functions which can perform pixel efficient adversarial attacks on both one-stage and two-stage object detectors.
	\item Our method can be easily extended to multi-models. We ensemble two off-the-shelf detectors and achieve considerable transferability in black-box attack settings.
	
\end{itemize}

\section{Method}

In this section, we explain our SAA method, which applying sparse adversarial patch on images to blind the detectors. The figure1 is an illustration of our SAA framework which ensemble multiple detectors to perform the evasion attack. To perform powerful adversarial attacks using limited number of pixels, we purposely design the position, shape and size of adversarial patch. For one-stage detector (YOLOv4 \cite{bochkovskiy2020yolov4}) and two-stage detector (FasterRCNN \cite{ren2016faster}), we design evasion loss function respectively to enhance our attack power. The rest of this section is organized as follows: section 2.1 illustrate our patch design, section 2.2 expound the loss function in SAA and section 2.3 shed light on optimizition details.

\subsection{Patch Design}

Our SAA can generate sparse adversarial noise. Concretely, we apply perturbation with bounded $l_{0}$ norm (no more than 2$\%$ of the image size) on the image. The perturbation is sparse in space but is efficient in fooling the detectors due to our purposely design of its spatial distribution.

In the training phase of detectors, the center of an object plays a vital role in the detection of this object. Take the one-stage detector YOLOv4 \cite{bochkovskiy2020yolov4} as an illustration, which will allocate anchor box that resemble ground truth mostly for an object. Even for two-stage detectors and anchor-free detectors, object centerness directly relates to the computation of intersection over union (IOU). Boxes that deviate object centerness too much are more likely to be erased during the non-maximum suppression (NMS) process.

Traditional adversarial patches with square or circle shape \cite{brown2017adversarial} \cite{liu2018dpatch} are too concentrated in a local area of the image. They have great physical realizability but are pixel inefficient in this task. We argue that a highly centralized structure limits the success rate of attacks especially when the $l_{0}$ norm constraint is too strong.

In view of those cases, we design a kind of cruciform patch with it's intersection locates at the centerness of object bounding box. Exploiting the long-span patch, we can mislead the results of detectors severely with very few pixels changed. 

Due to the difficulty of $l_{0}$ norm constrained optimization problem, we exploit a mask with values of 0 and 1 to decide the location of the patch, also to limit $l_{0}$ norm of the perturbation. We design the mask empirically based on the assumption that object centerness is a vulnerable area against adversarial attacks, combined with other conditions of the patch (shape, size) mentioned before. Hence the input of the detector is formulated as follows:
\begin{center}
	\begin{equation}
		Img = I \odot (1-M) + P \odot M	
	\end{equation}
\end{center}
Here $I$ denotes clean image, $P$ denotes adversarial patch, $M$ is the mask we design purposely and $\odot$ denotes element-wise product in this paper. As illustrated in figure 1, we include preprocesses of detectors into the forward phase of our attack in order to optimize the patch directly with gradient.

\subsection{Loss Function}
The goal of our SAA is erasing all the object detections in an image. And we define this type of attack on object detectors as evasion attack. The evasion attack is closely related to the definition of positive sample (foreground) and negtive sample (background) of the detector. We erase an object in an image by making it become a negative sample of the target detector. Consequently we design loss functions according to the definition of foreground and background of target detectors.

YOLOv4 \cite{bochkovskiy2020yolov4} is a one-stage object detector with excellent speed performance and considerable precision performance. YOLOv4 \cite{bochkovskiy2020yolov4} exploits a confidence branch to distinguish foreground and background. The bounding box with its confidence lower than a threshold in YOLOv4 \cite{bochkovskiy2020yolov4} will be recognized as background and will be discarded in the postprocess phase. Based on these principles, we design the evasion loss function of YOLOv4 \cite{bochkovskiy2020yolov4} as follows
\begin{center}
	\begin{equation}
		Loss_{YOLO}=\max \limits_{c \in C,b \in B}(conf(c,b))
	\end{equation}
\end{center}
Here $C$ denotes the set of all object categories, $B$ denotes the set of all bounding boxes and $conf$ is the object confidence in YOLOv4 \cite{bochkovskiy2020yolov4}. So the loss function extracts the maximum object confidence of an image. We minimize it to achieve our attack purpose.

FasterRCNN \cite{ren2016faster} is a typical two-stage object detector that firstly extracts bounding box proposals before the subsequent classification and regression processes. The definition of positive and negative samples is more complex in FasterRCNN \cite{ren2016faster} than that of YOLOv4 \cite{bochkovskiy2020yolov4}. However, we merely consider the inference phase of FasterRCNN \cite{ren2016faster} to simplify our attack. That is, we increase the softmax output probabililty of background while decrease that of any other categories (foreground objects). However, FasterRCNN \cite{ren2016faster} typically generates more than one hundred thousand region proposals by RPN, which is more than ten times that of YOLOv4 \cite{bochkovskiy2020yolov4}. The enormous number of region proposals makes it hard to erase all the objects with an extremely limited attack budget. So we make some compromises and design the total evasion loss function of FasterRCNN \cite{ren2016faster} as follows
\begin{center}
	\begin{equation}
		Loss_{FRCNN}=\alpha _{1} \cdot Loss_{1}+\alpha _{2} \cdot Loss_{2}
	\end{equation}
\end{center}
where $\alpha_{1}$ and $\alpha_{2}$ are hyperparameters, and we have

\begin{center}
	\begin{equation}
		\qquad \qquad Loss_{1}=\max \limits_{c \in C,b \in B}P(c,b)
	\end{equation}
	\begin{equation}
		\qquad \qquad Loss_{2}=\frac{1}{N} \sum \limits_{b \in B}\max \limits_{c \in C}P(c,b)
	\end{equation}
\end{center}
where $C$ denotes the set of all object categories, $B$ denotes the set of all bounding boxes predict by FasterRCNN \cite{ren2016faster} and $N$ is the number of elements of set $B$. We design $Loss_{1}$ to attack the bounding box with highest object probability, which is hard in experiments, hence $Loss_{2}$ is added to erase as many bounding boxes as possible. 

We ensemble the two detectors to train adversarial patches, thus the final loss function we use is as follows. Without deliberately balancing the weights of two terms in $Loss$, we can achieve high attack success rate.
\begin{center}
	\begin{equation}
		Loss=Loss_{YOLO}+Loss_{FRCNN}
	\end{equation}
\end{center}

\begin{figure}[htbp]
	\centering
	\vspace{0.1cm}
	\subfigure{
		\includegraphics[width=0.46\linewidth]{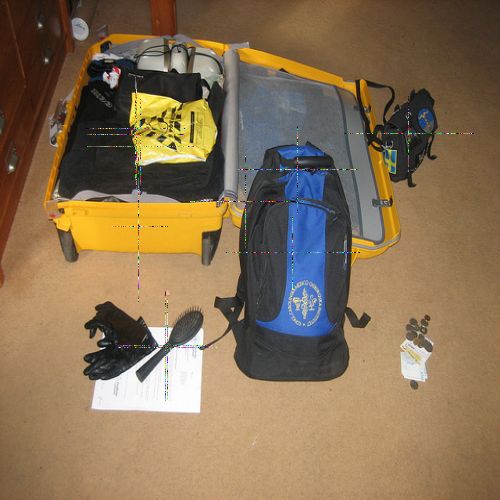}
	}
	\subfigure{
		\includegraphics[width=0.46\linewidth]{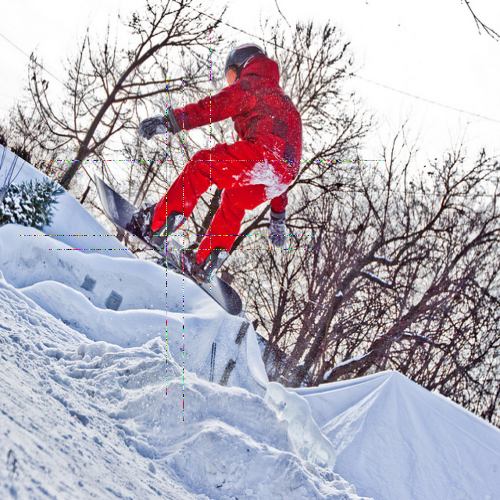}
	}
	\caption{Adversarial examples generated by SAA.}
\end{figure}

\subsection{Optimization Details}
With only 2$\%$ of pixels allowed to perturb, it's hard to generate adversarial examples with high attack success rate. We exploit essential tricks in the training process of adversarial patch to generate powerful adversarial examples.

To overcome the preprocess distortions, we include preprocess into the forward and backward phase of adversarial patch training, which including image resize, image normalization and other necessary processes. The strategy avoids unnecessary adversarial attack enhancements and thus accelerate our adversarial patch training process.

We exploit multi-scale steps to update adversarial patch and set all steps to integer multiples of $1/255$. We accelerate the training process tremendously by using large steps first. Also, it's a great way to avoid local minima in optimization. We decrease the update step gradually to refine our adversarial patch as the training goes on. It is worth mentioning that we set update steps to be special values (integer multiples of $1/255$) to filter out the quantification effect.

We also design multi thickness cruciform patch to adaptively attack images with different number of object detections. We believe thinner cruciform patch produce stronger attack effect for its adversarial effect on a wider range of image feature. Experiment results confirm its validity in practice.

In experiments, we conduct two phases attack to FasterRCNN \cite{ren2016faster}. At the first phase, we set $\alpha_{1}=1$ and $\alpha_{2}=0$ in $Loss_{FRCNN}$, aiming at suppress the bounding box with highest object probability. Due to rough patch location selection, we rarely succeed in the first phase to attack FasterRCNN \cite{ren2016faster}. And the second phase will be started once the first phase failed, we set $\alpha_{2}=1$ and $\alpha_{1}=0$ to attack as many as bounding boxes as possible. We don't set $\alpha_{1}\neq 0$ and $\alpha_{2}\neq 0$ together for better optimization of each term in $Loss_{FRCNN}$ at different phase.

\section{Experiments}
We select 1000 images from MSCOCO2017 dataset with all images resized to 500*500 size. And we choose YOLOv4 \cite{bochkovskiy2020yolov4} and FasterRCNN \cite{ren2016faster} as target models to conduct evasion attacks using our sparse adversarial patch. 

In order to comprehensively evaluate the performance of our attack method, we use evasion score as the evaluation indicator. For an image $x$ and its adversarial version $x'$,the evasion score in model $m$ is defined as follows
\begin{center}
	\begin{equation}
		S(x,x',m)=(2-\frac{\sum_{k}R_{k}}{5000})\cdot (1-\frac {min(B(x,m),B(x',m))}{B(x,m)})
	\end{equation}
\end{center}
where $R_{k}$ is $Kth$ connected domain of adversarial patch and $B(x,m)$ denotes the number of bounding boxes of image $x$ predicted by detector $m$. That means the fewer pixels you change and the more bounding boxes that disappear, the higher the evasion score. You get no more than 2000 evasion score in each detector.

\begin{table}[ht]
	\begin{center}
		\caption{Evasion Scores}
		
		\bigskip
		
		\begin{tabular}{|l|c|}
			\hline
			Detector(s) & Evasion Score \\ \hline
			YOLOv4 & 1610.03 \\
			FR-RES50 & 1174.21 \\ 
			Black-Box*2 & 355.69\\\hline
		\end{tabular}
	\end{center}
\end{table}

We ensemble two state-of-the-art object detectors YOLOv4 \cite{bochkovskiy2020yolov4} and FasterRCNN \cite{ren2016faster} to generate sparse adversarial examples using our SAA. The FasterRCNN we choose is ResNet-50 \cite{he2016deep} based and FPN \cite{lin2017feature} is introduced to optimize feature extraction. We also conduct experiments on two black-box detectors and achieve considerable transferability. The evasion scores of these detectors are listed in table 1. In white-box settings, our strategy achieve pretty high evasion scores of more than 1000 on both YOLOv4 \cite{bochkovskiy2020yolov4} and FasterRCNN \cite{ren2016faster}. Moreover we achieve 355.69 evasion score on two completely different black-box detectors, without any transferability enhancement strategy introduced.

\begin{figure}[htbp]
	\centering
	\vspace{0.1cm}
	\subfigure{
		\includegraphics[width=0.46\linewidth]{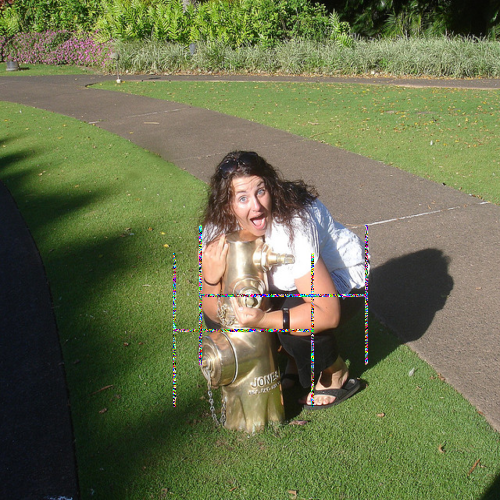}
	}
	\subfigure{
		\includegraphics[width=0.46\linewidth]{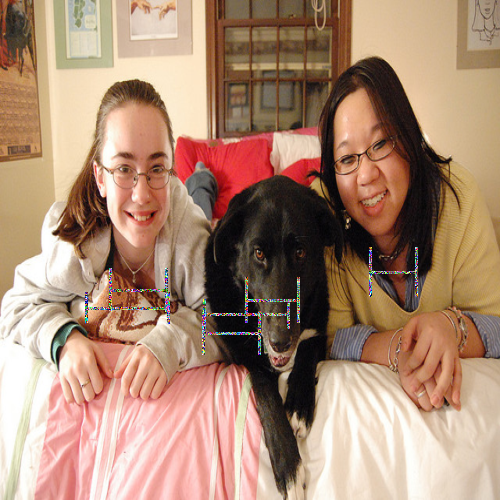}
	}
	\caption{Other shapes of adversarial patch.}
\end{figure}

We also make preliminary attempts to design different shape of the sparse patch, like what is shown in figure3, and get good results in some cases. We argue that better design of sparse adversarial patches is more effective in $l_{0}$ norm bounded settings. And we believe strategies like attention and saliency map can be exploited to improve our method.

\section{Conclusion}
In this work, we propose a sparse adversarial attack on object detectors. We design evasion loss functions to blind detectors with $l_{0}$ norm bounded perturbations. Our method achieve very high attack success rate on two state-of-art detectors and manifest considerable transforability even in black-box settings. Even so, we believe that our method can be further improved via selecting better locaions of the adversarial patch in the image.

\bibliographystyle{acm}
\bibliography{mybib}

\end{document}